\documentclass[10pt,twocolumn,letterpaper]{article}

\usepackage{cvpr}              

\definecolor{cvprblue}{rgb}{0.21,0.49,0.74}
\usepackage[pagebackref,breaklinks,colorlinks,allcolors=cvprblue]{hyperref}

\def\paperID{11106} 
\def\confName{CVPR}
\def\confYear{2025}

\title{DirectTriGS: Triplane-based Gaussian Splatting Field Representation for 3D Generation}

\author{{Xiaoliang Ju}, {Hongsheng Li}$^{1, 2}$\\
$^1$ CUHK MMLab   $\quad$  $^2$ CPII under InnoHK\\
{\tt\small akira@link.cuhk.edu.hk}
}

\usepackage{chapterbib}

\begin{document}
\maketitle
\begin{abstract}
We present DirectTriGS, a novel framework designed for 3D object generation with Gaussian Splatting (GS). GS-based rendering for 3D content has gained considerable attention recently. However, there has been limited exploration in directly generating 3D Gaussians compared to traditional generative modeling approaches. The main challenge lies in the complex data structure of GS represented by discrete point clouds with multiple channels.
To overcome this challenge, we propose employing the triplane representation, which allows us to represent Gaussian Splatting as an image-like continuous field. This representation effectively encodes both the geometry and texture information, enabling smooth transformation back to Gaussian point clouds and rendering into images by a TriRenderer, with only 2D supervisions. The proposed TriRenderer is fully differentiable, so that the rendering loss can supervise both texture and geometry encoding. Furthermore, the triplane representation can be compressed using a Variational Autoencoder (VAE), which can subsequently be utilized in latent diffusion to generate 3D objects.
The experiments demonstrate that the proposed generation framework can produce high-quality 3D object geometry and rendering results in the text-to-3D task.
  
\end{abstract}    
\section{Introduction}
 
Neural rendering has grown to a focal point in rendering techniques in recent years, as it achieves more realistic rendering effects by leveraging the great expressiveness of neural network.  The representatives are neural radience field~(NeRF)~\citep{mildenhall2021nerf} and the newly emerged Gaussian Splatting~(GS)~\citep{kerbl20233d}. However, when applied to the field of 3D generation, the slow rendering and training speed become a strong limitation for NeRF. While GS is more flexible with greater rendering efficiency and editability, few works have addressed the challenge of direct GS generation in 3D due to its complex data structure. In this paper, we propose a novel approach called DirectTriGS, which encodes GS as triplane representation, and introduce its corresponding renderer, TriRenderer. Subsequently, we apply latent diffusion on the Triplane representation to generate high-quality GS objects.

Gaussian splatting uses multi-channel point cloud of ``splats'' to describe the 3D contents. With a differentiable splats rasterizer, GS has fast rendering speed. However, the sparsity, multiple channels, and uneven distributed density of 3D GS bring great difficulty for 3D generation.
We propose to encode GS contents as multi-channel triplane representations, which have been shown to possess favorable properties for representing 3D geometry or  NeRF, as demonstrated in previous works ~\citet{wu2024blockfusion, shue20233d}. 
In our work, we leverage the triplane representation to encode both the geometry information and other GS attributes in two separate groups of channels. By training on a dataset of 3D objects, we obtain a shared TriRenderer that is capable of decoding different triplane representations to GS and then render it to image. 
TriRenderer is fully differentiable, enabling the use of only 2D rendering loss to supervise both the texture and geometry of 3D GS. The advantage of using Triplane to represent GS are two folds. First, it leads to high memory efficiency compared with dense voxels and is expressive enough for generating various 3D GS. Second, Triplane is more compatible with the convolution-based encoders compared with the original sparse GS point cloud, which require specifically designed networks on processing sparse point clouds.

We follow stable diffusion~\citep{rombach2022high} to train DirectTriGS on the proposed triplane representation.
Specifically, a VAE is designed to further convert the Triplanes into latent code. We use two separate decoders to decouple the decoding of geometry and GS attributes. Second, we roll out the triplane latent to an expanded multi-channel image, then exploit staged latent diffusion to do generation. Two-stage diffusion is employed to generate the geometry and the corresponding GS appearance. Finally, score distillation sampling~(SDS)~\citep{poole2022dreamfusion} is adopted as a optional post-processing to refine or restyle the generated 3D objects.

Our contribution can summarized as follows. 1) We propose a Triplane representation for direct 3D GS generation, which is memory efficient to derive GS point clouds with vivid rendering. 2) We
design a fully differentiable TriRenderer to enable the end-to-end training of triplane representations, with only 2D supervisions. 3) We develop a Triplane-based GS generation framework DirectTriGS that incorporates a specially designed VAE, latent diffusion module, and a SDS based refiner.  4) Experiments demonstrate our method produces competitive performance in both 3D geometry and multi-view rendering quality in text-to-3D tasks.

\section{Related Works}

\noindent \textbf{3D Object Generation.}
1)Multi-view images-to-3D. The first route generates multi-view colored 3D image and then reconstruct the 3D shape and project image to texture, or use score distillation sampling to build 3D object from images produced by 2D diffusion, such as \cite{poole2022dreamfusion, shi2023mvdream, hollein2023text2room, liu2023zero, chung2023luciddreamer, tang2023dreamgaussian}. Since these approaches are actually 2D generator,  it is difficult to maintain 3D consistency, which may lead to Janus problem.  2)Image-to-3D.
The second route attempt to retrieve 3D geometry from a single input image, such as \cite{zou2023triplane, liu2024one, long2024wonder3d}. With the certain one image as input, the users may be more concerned with the reconstruction quality.
3)Text-to-3D. The third group mainly operate in 3D, which can be clustered by different 3D representation.
For example, \cite{gupta20233dgen,gao2022get3d} are able to generate high-quality 3D textured meshes using differentiable rendering;
\cite{nichol2022point,wu2023sketch} generate colored pointcloud;
\cite{chen2023single,metzer2023latent,lan2025ln3diff, hong20243dtopia} generate NeRF volumes;  \cite{ju2023diffindscene,wu2024blockfusion}generate SDF volumes. 

\noindent {\textit{Remark.}} The input conditions of these three generation methods are gradually becoming abstract, corresponding to increasing difficulty of the generation task, commonly resulting in a gradual decline in the generation quality.

\noindent \textbf{Gaussian Splatting.}
Gaussian Splatting~\citep{kerbl20233d} exploits multi-channel pointcloud of splats to describe the 3D content, which allows real-time rendering of photorealistic scenes. 
Recently, there emerges so many research focusing on its reconstruction quality enhancement and algorithm optimization~\citep{yu2023mip,cheng2024gaussianpro,fan2023lightgaussian,lu2023scaffold, huang20242d}. 
Another related active area includes 4D dynamic GS modeling~\citep{wu20234d,yang2023real,chen2023periodic} , scene editing~\citep{fang2023gaussianeditor,chen2023gaussianeditor} and its applications like SLAM~\citep{matsuki2023gaussian,yan2023gs}.

As for GS generation, some works exploit multi-view image based route such as \cite{chung2023luciddreamer,tang2023dreamgaussian}, where the image generator output image from required views to enhance the GS rendering in a recursive manner. Rare works explore the GS generation directly in 3D space, such as
\cite{zou2023triplane} generates a point cloud using a Transformer structure with 2D image as input, and build a mapping of image to tri-plane encoding of GS attribute; GaussianCube~\citep{zhang2024gaussiancube} proposes to use optimal transport to model the 3D GS for text-to-3D generation; \cite{jiang2024brightdreamer} combine explicit anchor points with implicit triplane to directly generate GS points.

\section{Revisiting Gaussian Splatting}
Gaussian Splatting~\cite{kerbl20233d} uses a set of Gaussian points to describe a 3D object.
The Gaussian points are defined by a full 3D covariance matrix $\mathbf{\Sigma}$ in a world space as $G(\mathbf{x})=e^{-\frac{1}{2}\mathbf{x}^T\mathbf{\Sigma}^{-1}\mathbf{x}}$, centered at point~(mean) $\mathbf{p}\in\mathbb{R}^3$. Each point is with a opacity scalar $\alpha\in\left[0, 1\right]$ for blending and a series of Spherical Harmonics (SH) coefficients to correctly capture the view-dependent appearance of the scene. The number of SH coefficients are $3 \times(n+1)^2$, where $n$ denotes the SH order, and higher $n$ corresponds to more accurate view-dependent appearance.

During rendering, the Gaussians are project to 2D given a viewing transformation $\mathbf{W}$. The covariance is transformed as $\mathbf{\Sigma}'=\mathbf{JW\Sigma} \mathbf{W}^T\mathbf{J}^T$, where $\mathbf{J}$ is the Jacobian of the affine approximation of the projective transformation. To ensure $\mathbf{\Sigma}$ to be semi-positive in the whole training process,  $\mathbf{\Sigma}$ can be represented by a Cholesky decomposition as $\mathbf{\Sigma}=\mathbf{RSS}^T\mathbf{R}^T$, represented by a tuple $s=(s_1,s_2,s_3)$ for scaling and a unnormalized quaternion $\mathbf{q}\in\mathbb{R}^4$ for rotation.
In summary, each Gaussian point are a union set of position $\mathbf{p}$ and its GS attributes:
\begin{equation}
    \label{eq:gs}
    \mathcal{G}_{\mathbf{p}}:=\{s, \mathbf{q}, \mathrm{SH}, \alpha \}.
\end{equation}
GS pointcloud can be efficiently rendered by a rasterization-based splatting renderer~\citep{kerbl20233d}, which is fully differentiable. 

\section{Methodology}
\begin{figure*}[t]
    \centering
    \includegraphics[trim={3.6cm 7.5cm 4.1cm 3.0cm},clip, width=0.9\linewidth]{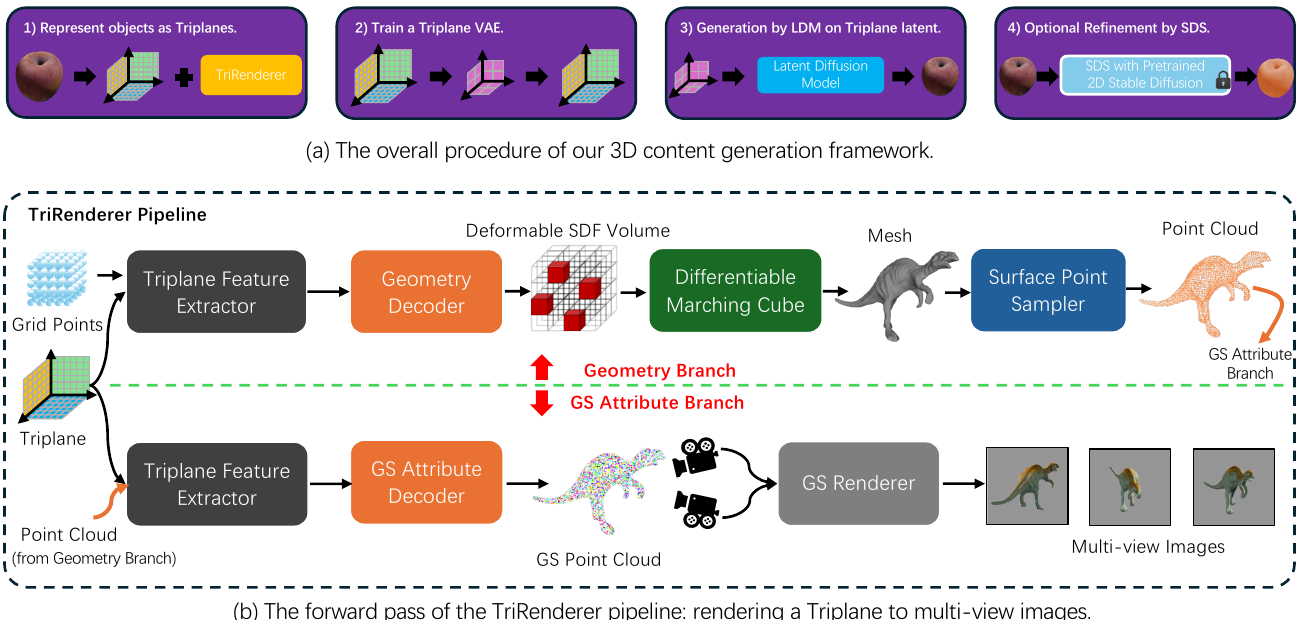}
    \caption{The overview of the proposed 3D GS generation framework and the core component TriRenderer. We use triplane to represent 3D objects, and design a TriRenderer to decode triplane to GS pointcloud and render multi-view images.}
    \label{fig:model}
\end{figure*}

Our motivation is to directly generate 3D GS with fast speed by utilizing the proposed triplane-based GS representation. While 2D-lifting methods like SDS-based approaches yield intricate results, they are time-intensive and susceptible to Janus problem. Direct generation, notably with fast sampling, can complete 3D generation in under a minute, contrasting with SDS methods that take over 10 minutes or longer, such as \cite{chen2024text}\cite{wang2024prolificdreamer}, posing challenges for users. Therefore, direct 3D generation is more practical, with 2D-lifting reserved for enhancing texture details when needed.

The procedures of our generation framework can be divide into 4 stages as shown in Fig.\ref{fig:model} (a). 
Firstly, we encode the 3D objects into triplanes using the TriRenderer. The triplane encoding and TriRenderer training can be concurrently executed with only the multi-view RGBA image and camera poses as supervision. 
Secondly, a VAE is trained to compress the triplanes to latent space, enabling effective capture of high-level information for subsequent modules.
Thirdly, a diffusion model is trained on the triplane latent code. 
During inference, the diffusion model produces the latent code, which can be decoded back to its corresponding triplane by the VAE decoder. Subsequently, the TriRenderer decodes the triplane into the standard GS point cloud.  
Finally, an SDS-based refiner can be utilized as an optional post-processing step to enhance or restyle the generated 3D object. Leveraging the pretrained 2D diffusion model within the SDS, the generated GS can be further improved to achieve a more detailed appearance.

\subsection{Triplane-based GS Field}
\label{sec:triplane}

We employ triplanes to convert the discrete multi-channel GS points into a continuous field, tailored for more efficient encoding in the subsequent generative model.
Triplanes are 3 planes formed by every 2 axes of $x,y,z$ axes in 3D space, where every 3D point $\mathbf{p}$ can query these planes to get corresponding features $F_{\mathbf{p}}=(F_{xy}, F_{xz}, F_{yz})$ using orthogonal projection. 
In real implementation, triplane is a tensor with the size of $3\times H\times  H \times C$, where $H$ denote the resolution along every axis and $C$ is the feature channel. The query of point with arbitrary continuous coordinate should be a bi-linear intepolation in the triplane grids. 
Considering GS is a set of points with multiple channels of attributes, it is natural to use triplane for GS encoding. If the triplane feature and GS attributes of a point can be transformed to each other, the sparse GS pointcloud of every object can be represented as continuous triplane GS field for further encoding. We use separate channels to encode 3D geometry and other GS attributes as described in Eq.~\ref{eq:gs}, which is an experimental setting for better convergence in the training.

\subsection{TriRenderer} 
\label{sec:trirenderer}
The TriRenderer serves as the crucial component for converting a 3D object into a triplane-based GS field. It acts as a fully differentiable bridge connecting the GS field with rendering, enabling the optimization from 2D images to geometry and appearance encoding.

As depicted in Fig. \ref{fig:model}~(b), the TriRenderer comprises a geometry branch and a GS attribute branch, each equipped with independent decoders to decode the triplane.
The geometry branch is responsible for retrieve the geometry as triangular mesh from triplane, with a surface sampler for GS pointcloud sampling. Then GS attribute branch uses the obtained pointcloud to query the triplane to obtain the GS attribute corresponding to each point. In this way, the GS pointcloud in the original format is retrieved, and can be rendered using the original GS renderer. It is worth mentioning that all triplanes of different object share a common TriRenderer, which ensures the features on different triplanes subject to a similar distribution.

\noindent \textbf{Geometry Branch.} 
We use signed distance function (SDF) to represent the geometry, and allow every vertex deform as \cite{shen2021dmtet},
so that the geometry branch decodes $F_{p}$ to a SDF volume and its vertice deformation. 
By query all grid coordinates of the designated resolution $L\times L\times L$,  the deformable SDF volume is reconstructed.  Then we exploit a differentiable Marching Cube algorithm FlexiCubes~\citep{shen2023flexicubes} to extract triangular mesh from the SDF volume. 

\noindent \textbf{Surface Sampling.} Considering that GS pointcloud is generally gathered on the surface of objects, we randomly sample GS points on the faces of triangular mesh with barycentric coordinates. We expect that every splat is flat and with a normal consistent with its source triangular face.
Assume a face is formed by vertices $<v_a, v_b, v_c>$, and vector $\mathbf{v}_1=(v_a - v_b)$, $\mathbf{v}_2=(v_b - v_c)$, the face normal can be calculated as 
$\mathbf{n}=\mathbf{v}_1\times \mathbf{v}_2$. Then the rotation matrix of every splat drawn from this face can be formulated as 
\begin{equation}
    \mathbf{R} = \left[\mathbf{v}_1, \mathbf{n}\times\mathbf{v}_1, \mathbf{n} \right],
\end{equation}
which can be transform to $\mathbf{q}$ in Eq.\ref{eq:gs} by standard matrix-to-quaternion algorithm. To make the splats flat, $s_3$ in Eq.\ref{eq:gs} is fixed to an infinite small value.

To further reduce the computational load in the subsequent GS rendering process, only faces oriented toward the camera are sampled. These face indices can be obtained through fast rasterization of the mesh faces. Note this rasterization is not required to be differentiable, and we use the library Nvdiffrast~\citep{Laine2020diffrast} for implementation.

\noindent \textbf{GS Attribute Branch.} 
By querying the triplane GS channels with the sampled pointcloud, we can obtain the GS feature and use the GS attribute branch to transform them to the rest GS attributes. Then the GS renderer can render GS pointcloud to multi-view images. Since different GS attributes have different numerical scales and distributions, we customize individual headers to decode them respectively, which is similar to \cite{zou2023triplane}.

\noindent \textbf{Training.}
 The training consists of 2 stages. In the first stage, a small batch of data is used to train the triplanes and the TriRenderer together. The second stage involves the whole dataset to train all triplanes with the parameters of TriRenderer frozen. In this way, we can handle large dataset by distributed training in the second stage, and the same TriRenderer can be shared by all triplanes. There is no need to pre-train the GS pointcloud using original GS training as \cite{kerbl20233d}. 

 The only required supervision data is $N$ multi-view images $\{I_i\}_{i=1}^N$ with camera poses, and the training losses consists of the rendering loss $L_{\mathrm{render}}$ and the geometric regularization $L_{\mathrm{geo}}$. The rendering loss is a weighted sum as 

\begin{equation}
    \label{eq:l_render}
    L_{\mathrm{render}} = w_{1}L_{\mathrm {alpha}}+w_{2}L_{\mathrm {rgb}}+w_{3}L_{\mathrm {pips}},
\end{equation}
 where $L_{\mathrm {alpha}}$ is a $L_1$ loss on the alpha map, namely the silhouette image loss; $L_{\mathrm {rgb}}$ is a combination of $L_1$ pixel loss and SSIM loss~\citep{wang2004image} between rendered image $\hat{I_i}$ and ground truth ${I_i}$:
\begin{equation}
\label{eq:rgb_loss}
    L_{\mathrm{rgb}} = \sum_{i=1}^N(1-\beta)\|I_i-\hat{I_i}\|_1 + \beta {\mathrm {SSIM}}(I_i,\hat{I_i}),
\end{equation}
and $L_{\mathrm {pips}}$ is the perceptual loss~\citep{johnson2016perceptual}. $w_1$, $w_2$, $w_3$ and $\beta$ are all weighting factors determined by experiments.

The geometric regularization loss on the SDF volume $V$ is defined as 
\begin{equation}
    \label{eq:l_geo}
    L_{\mathrm{geo}} = \gamma_1L_{\mathrm{dev}} + \gamma_2L_{\mathrm{weighting}} + \gamma_3L_{\mathrm{CE}} + \gamma_4L_{\mathrm{sign}} ,
\end{equation}
where the $L_{\mathrm{dev}}$ and $L_{\mathrm{weighting}}$ are defined by Flexicubes~\citep{shen2023flexicubes} to regularize the extracted connectivity and the weighting scale of SDF vertices. To penalize the sign changes on all grid edges,
we follow \cite{munkberg2022extracting} to define $L_{\mathrm{CE}}$ as
\begin{equation}
    L_{\mathrm{CE}} = \sum_{(s_a, s_b)\in\epsilon} \mathrm{CE}(\sigma(s_a), \mathrm{sign}(s_b)),
\end{equation}
 where $\mathrm{CE}$ denotes cross entropy; $\epsilon$ is the set of all edges connecting vertices with different signs; and $\sigma$ is the sigmoid function.
 Finally, $L_{\mathrm{sign}}$ is designed to prevent the SDF volume from being trapped in an empty shape, i.e. a fully positive or fully negative SDF volume, 
\begin{equation}
\label{eq:sign}
    L_{\mathrm{sign}} = \delta(V) M(|V|),
\end{equation}
where $\delta(V)=1$ if $V$ is empty, otherwise $\delta(V)=0$. $M(|V|)$ denotes the mean of the absolute value of $V$.

\begin{figure*}[t]
    \centering
    \includegraphics[trim={1cm 7cm 3.5cm 8cm},clip, width=0.9\linewidth]{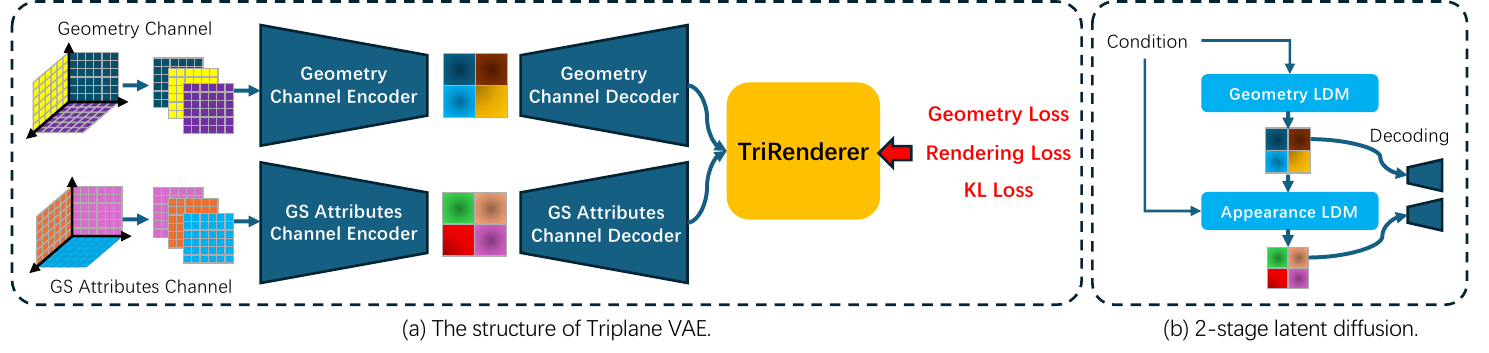}
    \caption{Triplane VAE and 2-stage diffusion.}
    \label{fig:vae_diff_structure}
\end{figure*}

\subsection{Triplane VAE}

We employ an UNet-like structure of VAE to compress the triplane to latent space. The training pipeline is demonstrated in Fig.~\ref{fig:vae_diff_structure}~(a). Considering that the 3 planes of triplanes encodes features from 3 orthogonal view direction, these 3 planes should be homogeneous data, so that We reshape the triplanes of batchsize $B$ to a new batch as $B\times 3\times  H \times H\times C $ to $3B\times H \times H \times C$ for the VAE training. We use decoupled encoders and decoders for geometry and  GS channels of triplanes, for experiments prove that a mixed encoding may lead to blur of the rendering results. After the triplane is reconstructed, the TriRenderer trained in Section.\ref{sec:triplane} can be used to retrieve the GS pointclouds and render it to images. 

In the training process, same loss functions  as   Eq.~\ref{eq:l_render}  and Eq.~\ref{eq:l_geo} are used, along with an $L_1$ loss between the input and reconstructed triplanes, denoted as $L_{\mathrm{Tri}}$. 
Additionally, a Kullback–Leibler divergence loss $L_{\mathrm{KL}}$ is included to ensure that the latent variables do not deviate significantly from a normal distribution. The total losses is summarized as
\begin{equation}
    L_{\mathrm{VAE}} = L_{\mathrm{Tri}} + L_{\mathrm{Geo}} + L_{\mathrm{Render}} + \gamma L_{\mathrm{KL}}, 
\end{equation}
where $\gamma$ is a small weight.

\subsection{Two-stage LDM for Text-to-3D Generation}
\label{sec:ldm}

Taking into account the significant relevance of texture appearance to the underlying geometry, we proposed to utilize staged diffusion to generate geometry and GS attribute successively as shown in Fig.~\ref{fig:vae_diff_structure}~(b). It is easy to implement it because the latent codes for geometry and other GS attributes are totally decoupled by VAE, as mentioned earlier .
In this way, the geometry code can be a new condition for the second stage.

We follow DDPM~\cite{ho2020denoising, karras2022elucidating} to implement the latent diffusion conditioned on the text description. To better capture the relation between different planes, we roll out the triplane latent of $B\times 3\times  h \times h\times c$ to a image-like $B\times 3h \times h\times c $ as the input of diffusion model, and the generated result will be transformed back to the original shape for TriRenderer decoding.

\subsection{SDS-based Texture Booster}
\label{method:sds}
\begin{figure*}[t]
    \centering
    \includegraphics[trim={1cm 16.5cm 13.8cm 3cm},clip, width=0.85\linewidth]{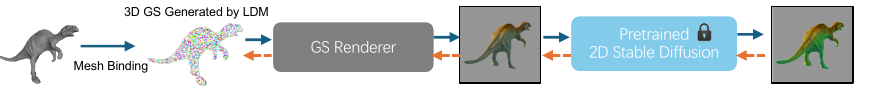}
    \caption{Score distillation process(SDS) for optional texture refinement.}
    \label{fig:sds_method}
\end{figure*}
To refine or restyle the generated GS pointcloud,  we utilize a pretrained 2D diffusion model for the SDS training process~\citep{poole2022dreamfusion} as shown in Fig.~\ref{fig:sds_method}. 
As the generated 3D GS is constrained by the mesh reconstructed from SDF, detailed in Section \ref{sec:trirenderer}, we maintain the GS splats' adherence as previously established. However, we enable the mesh vertices to shift within a limited radius to enhance the stability of the SDS process. Additionally, throughout the entire SDS procedure, there is no need for densification or pruning operations. This ensures that the number of points can be controlled to align with the number of mesh faces.

With the RGB image rendered from the initial GS pointcloud, the 2D diffusion model can produce images with better 2D/3D consistence in different views, so that the optimization process can converge rapidly without Janus problem.

\section{Experiment}
The experiments are conducted step-by-step according to the generation pipeline proposed as Fig.\ref{fig:model}~(a).  We use OmniObject3D~\citep{wu2023omniobject3d} dataset for toy experiments, and Objaverse~\citep{deitke2023objaverse} dataset for main evaluation. 
First, we sample 200K object to train their cooresponding triplanes and TriRenderer. Then we train the VAE and  diffusion model for 3D GS generation. Finally, we exploit SDS to boost the texture quality for unsatisfactory generated objects.
The procedure of data pre-processing and the implementation details are included in the supplementary materials.

The experimental results primarily consist of the following components:
1) For examining the proposed triplane modeling for GS generation, we focus on the text-to-3D task and showcase the direct GS generation outcomes qualitatively and quantitatively. These results are compared with Shap-E~\citep{jun2023shap}, Direct3D~\cite{liu2024direct}, and the most recent GaussianCube~\citep{zhang2024gaussiancube}.
2) In comparison to 2D-lifting GS generation methods, we present the performance of SDS refinement on unsatisfying samples, with GSGEN~\citep{chen2024text} and GaussianDreamer~\citep{yi2024gaussiandreamer} as baselines.

\subsection{Triplane Fitting and VAE Reconstruction} 

\noindent \textbf{Triplane Fitting.} We adopt a resolution of $3\times128\times128\times 16$ for all triplane fitting, with a resolution of $64\times64\times 64$ for the deformable SDF grids. We randomly sample 1000 objects for the shared TriRenderer training, and train all residual objects distributively. Every triplane is initialized with Gaussian noise before training.
Under such settings, every triplane takes less than 30 seconds for geometry and GS appearance reconstruction.

\noindent \textbf{VAE Reconstruction.}We use a down-sample factor of 4 to compress the triplanes to latent space. 

We visualize the triplane and the VAE reconstruction results of several data samples for a simple check, please refer to the supplementary materials.

\subsection{Direct Text-to-3D GS Generation by LDM} 

In this section, we present the results of our 3D GS object generation without the SDS refinement, both quantitatively and qualitatively. 
Due to the current scarcity of work targeting direct 3DGS generation, we selected two state-of-the-art text-to-3D works based on other 3D representations(NeRF) Shap-E~\citep{jun2023shap} and Direct3D~\citep{liu2024direct} for comparison. For fairness, these selected  methods also perform direct generation in the 3D domain {\bf without} any 2D prior knowledge or refinement based on 2D diffusion. Additionally, for the limited space, we provide more generation results and comparison with other methods such as 
3DTopia~\citep{hong20243dtopia},
GaussianCube~\citep{zhang2024gaussiancube}
, and LN3Diff~\citep{lan2025ln3diff} 
in the supplementary materials.

\begin{figure*}
    \centering
    \includegraphics[trim={1cm 2.5cm 16.3cm 2.5cm},clip, width=0.95\linewidth]{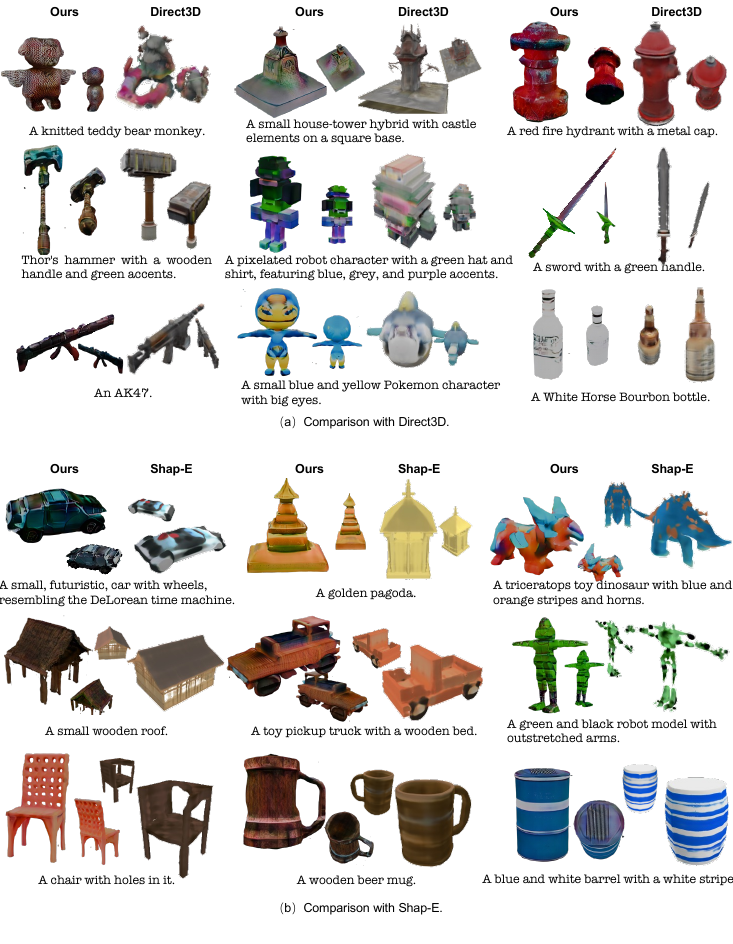}
    \caption{Comparison of generated 3D objects ({\bf Raw results without any refinement such as SDS}). }
    \label{fig:gen1}
\end{figure*}

\noindent \textbf{Qualitative Results.}
The generation results from different methods are listed in Fig.~\ref{fig:gen1}. Every generated sample is rendered from different views accompanied by the provided text prompt captioned beneath the images. Our method showcase enhanced proficiency in both geometry and rendering quality, resulting in sharper and clearer outputs, which can be further verified in the subsequent quantitative evaluation. For more generated samples, please check the supplementary materials.

\noindent \textbf{Quantitative Results.} We use CLIP score to evaluate the text-to-3d consistency, and an user study is conducted to evaluate the generation results from various aspect such as geometry, texture, realistic rendering and the consistency with given prompt. 
49 users participated in the user study to score the over 50 3D samples from 1 to 5 points, and the average results are shown in Tab.~\ref{tab:user_study1}. As for the CLIP score, the open-source repository \texttt{t2vmetrics}~\citep{lin2024evaluating} is used to calculate the CLIP score on two versions of ViT models, and the results are demonstrated in Tab.~\ref{tab:clip_score}. Both the CLIP score and the user study indicate that the proposed method produces better performance.

\begin{table}[]
\centering
\resizebox{\columnwidth}{!}{
\begin{tabular}{lccccc}
\toprule
         & Geometry$\uparrow$  & Texture$\uparrow$  & Realism$\uparrow$ &  Consistency$\uparrow$ & Overall$\uparrow$ \\
         \midrule
Shap-E   &3.080&3.019                  &    2.947                &      3.121               &    3.042                        \\
Direct3D & 3.222                 & 3.150                   &         3.118            &         3.242           &    3.183     \\
Ours     &   3.456               &   3.383                 &     3.332                &        3.520            &  3.423 \\
\bottomrule
\end{tabular}
}
\caption{User study on generated 3D objects ({\bf Our results are raw  outputs from LDM without SDS refinement}).}
\label{tab:user_study1}
\end{table}

\begin{table}[]
\centering
\resizebox{0.95\columnwidth}{!}{
\begin{tabular}{lccc}
\toprule

                    & Shap-E & Direct3D & Ours   \\
                     \midrule
openai:ViT-L-14     & 0.2398 & 0.2152   & 0.2456 \\
openai:ViT-L-14-336 & 0.2426 & 0.2220   & 0.2462 \\
\bottomrule
\end{tabular}
}
\caption{CLIP score($\uparrow$) for evaluation of similarity between rendered images and given text prompt(({\bf Our results are raw  outputs from LDM without SDS refinement})).}
\label{tab:clip_score}
\end{table}

\subsection{Optional Texture Boosting by SDS} 

\begin{figure*}
    \centering
    \includegraphics[trim={1cm 2.9cm 15cm 2.5cm},clip, width=0.83\linewidth]{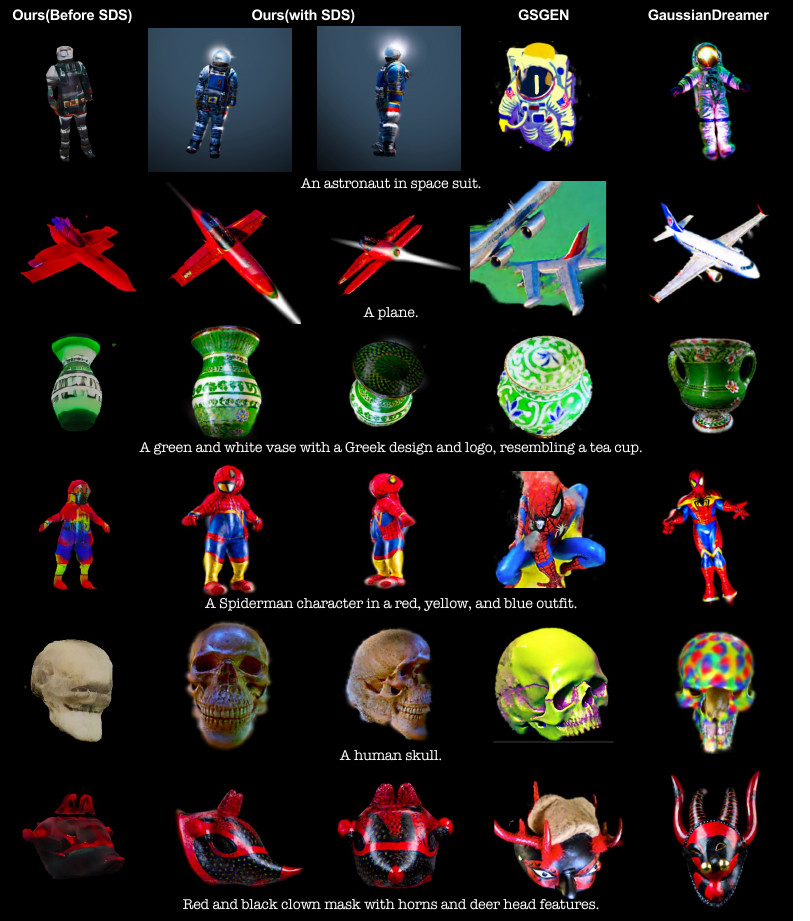}
    \caption{Refinement of unsatisfactory results by SDS, compared with pure SDS-based methods GSGEN and GaussianDreamer.}
    \label{fig:gen2}
\end{figure*}

To compare with other SDS-based 3D GS generation methods, we implement a version of SDS for our mesh-binding GS representation as described in Section \ref{method:sds}. GSGEN~\citep{chen2024text} and GaussianDreamer~\citep{yi2024gaussiandreamer} are selected as our baselines.
The generated samples, as illustrated in Fig.~\ref{fig:gen2}, highlight that with the SDS refiner, our DirectTriGS can deliver competitive outcomes comparable to the state-of-the-art 2D-lifting methods.

Furthermore, we conducted a user study for quantitative evaluation. 40 users were tasked with ranking 15 samples from different methods based on two criteria: realism and level of detail. The results indicate that our approach marginally outperforms the other baseline methods.

\begin{table}[]
\centering
\resizebox{0.95\columnwidth}{!}{
\begin{tabular}{lccc}
\toprule
         & More Realistic (\%) & More Detailed (\%) & Overall (\%) \\
         \midrule
GaussianDreamer   &35.0&        50.0                                 &     42.5         \\
GSGEN &     17.5                &    5.0                &11.3              \\
Ours     &47.5&          45.0           &      46.3                  \\
\bottomrule
\end{tabular}
}
\caption{User study on generated 3D objects (with SDS refinement).}
\end{table}

\subsection{Ablation Study} 
\begin{figure*}[]
\centering
\begin{minipage}[]{0.216\linewidth}
\centering
\includegraphics[trim={1.cm 9cm 18cm 1cm},clip, width=0.8\linewidth]{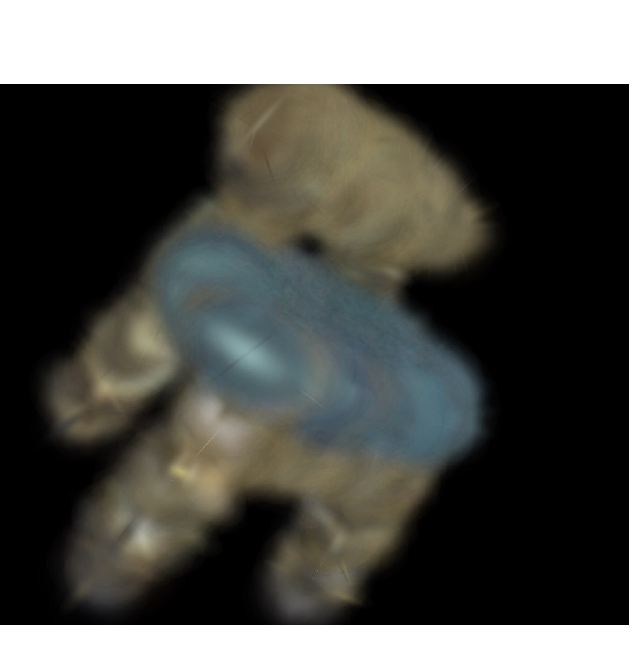}
\caption{Failure of GS attribute generation.}
\label{fig:ablation_3ddiff}
\end{minipage}
\hspace{0.3cm}
\begin{minipage}[]{0.568\linewidth}
\centering
\includegraphics[trim={5.cm 7cm 4cm 8cm},clip, width=\linewidth]{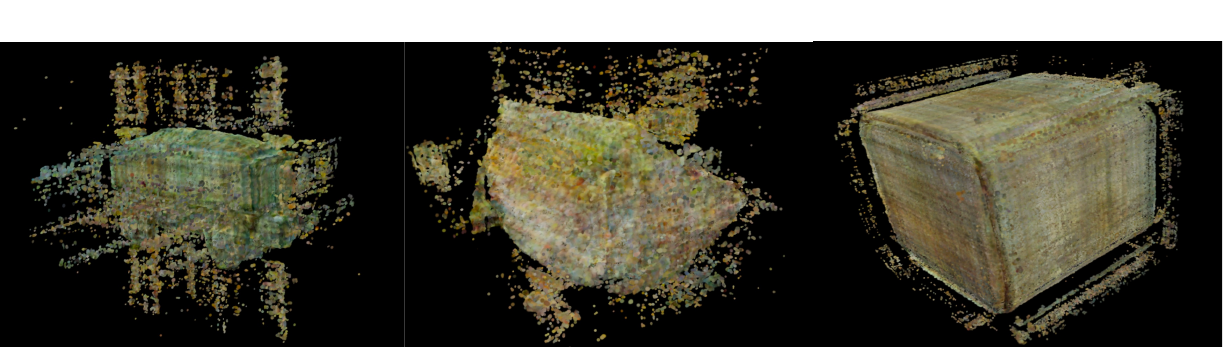}
\caption{Noisy results of direct diffusion on triplane. The condition prompt are "train", "watermelon" and "box".}
\label{fig:ablation_tridiff}
\end{minipage}
\end{figure*}

\noindent \textbf{3D Diffusion for voxel-based Gaussian Splatting.} We attempt to do GS attribute generation conditioned on given voxel occupancy via 3D diffusion model. However, even a toy experiment of over-fitting one single object failed, which may attributes to the complex multiple channels with different distributions especially the non-Euclidean ones such as quaternion. As shown in Fig.\ref{fig:ablation_3ddiff}, the diffusion model learns to generate color but fails to generate splat scaling, opacity and orientation.

\noindent \textbf{Triplane Diffusion without VAE.} Since VAE inherently involves a reconstruction loss, we attempt to use direct diffusion on the triplane space for generation. The experiment results shows that such direct diffusion may cause serious noise on decoded geometry. Randomly generated samples are visualized in Fig.\ref{fig:ablation_tridiff}. A possible reason is that the multi-channels of triplane contains considerable redundancy or noise, which is difficult to be captured or filtered by the diffusion model.

\noindent \textbf{Inference Efficiency.} The inference efficiency is listed in Table \ref{tab:inference_eff}. This experiment is conducted on the platform equipped with RTX3090 GPU with 24GB memory.
\begin{table}[]
\centering

\resizebox{\columnwidth}{!}{
\begin{tabular}{lcccc}
\toprule
           & TFLOPs  & Paras.(M) & GPU Mem.(GB) & Running Time (Sec.)       \\ \hline
Generation     & 0.00823 & 56.65     & 2.94    &  16.2\\
\bottomrule
\end{tabular}
}
\caption{Inference efficiency of the generation, with the batch size is 1.}
\label{tab:inference_eff}
\end{table}
\section{Conclusion}
We have presented a novel framework DirectTriGS for Gaussian Splatting Field generation. DirectTriGS mainly consists of 3 parts: 1) a light-weight triplane representation for 3D object with the format of Gaussian Splatting; 2) a fully differentiable TriRenderer which can decode triplane to orginal GS and render it to multi-view images; 3) the triplane VAE and staged diffusion model for the whole generation process. By utilizing our DirectTriGS, the intricate GS data can be generated directly and efficiently. 4) Additionally, we incorporate a SDS refiner to further improve the texture and details of generated objects.

\newpage
{
    \small
    \bibliographystyle{ieeenat_fullname}
    \bibliography{main}
}

\clearpage
\setcounter{page}{1}
\maketitlesupplementary

\section{Dataset Information.}
Objaverse~\cite{deitke2023objaverse} is the main dataset for our experiment, which contains over 800K 3D objects. As the rendering process on such a massive dataset is very time-consuming, we adopt the pre-processed version sourced from the repository  of \cite{qiu2023richdreamer}, which pre-filters over 260K samples. In this processed dataset, every object is normalized to the voxel range of 
$[\pm0.5, \pm0.5, \pm0.5]$, and rendered to RGBA images in a resolution of $512*512*4$, with 40 views in total.
Our training data only comprises multi-view images and their corresponding camera poses, without any kind of original 3D data.

\section{Implementation Details.}
\noindent \textbf{Triplane.}
The triplane resolution is configured as $3\times128\times128\times16$, where 16 represents the channels within each grid. The first half of the channels is designated for encoding geometry information, while the remaining half is allocated for encoding GS appearance details. Each triplane is initialized to random Gaussian noise with a standard deviation of 0.01. This random initialization allows the triplane to be decoded into random SDF values, subsequently leading to the generation of diverse fragmented mesh faces.
Upon rasterization of these faces onto the screen, the geometry loss facilitates swift removal of undesired faces.
During our experiments, we observed that this initialization method enables faster convergence compared to zero initialization. 

As for loss configuration, we configure $w_1=5.0$, $w_2=1.0$, $w_3=1.2$, $\beta=0.2$, $\gamma_1=0.2$, $\gamma_2=0.1$, $\gamma_3=0.01$, $\gamma_4=1.0$ by experiments, corresponding to the loss function described in Eq.~\ref{eq:l_render}, Eq.~\ref{eq:rgb_loss} and Eq.~\ref{eq:l_geo}.

\noindent \textbf{TriRenderer.}
As for the TriRenderer instroduced in Fig.\ref{fig:model}, both the geometry decoder and the GS attribute decoder inside it are composed of linear blocks. In the GS attribute decoder, there are 3 headers for GS splats  scaling, opacity and SH prediction, and the rotation is fixed by the mesh face normal as introduced in Section \ref{sec:trirenderer}. All the GS attribute headers are linear layers. We set SH degree to 1 in all experiments, which is enough to obtain satisfying results on Objaverse.

\section{Simple Check of Triplane and VAE.}

\begin{figure}[h]
    \centering
    \includegraphics[trim={2cm 4cm 20cm 6cm},clip, width=\linewidth]{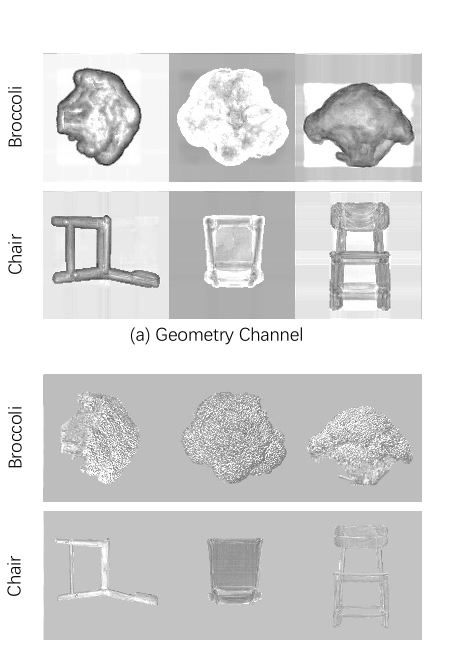}
    \caption{Channel visualization of sample triplanes. }
    \label{fig:tri_vis}
\end{figure}

\begin{figure*}[h!]
    \centering
    \includegraphics[trim={1.5cm 11cm 11cm 7cm},clip, width=0.8\linewidth]{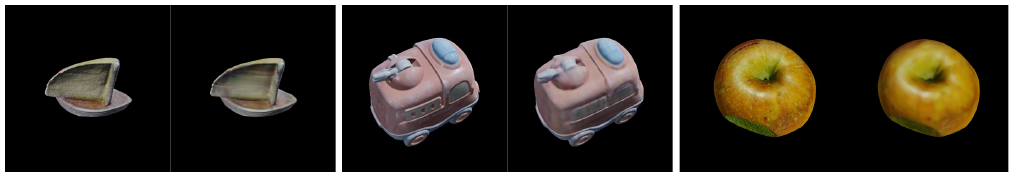}
    \caption{Triplane reconstructed by VAE. Left: ground truth. Right: reconstruction.}
    \label{fig:vaeloss}
\end{figure*}

To better investigate whether it is reasonable to encode triplane using convolution-based methods, we simply scale the channel value of trained triplanes to pixel range and visualize them as shown in Fig.~\ref{fig:tri_vis}, where clear shapes from 3 different views can be observed. As for the VAE reconstruction, a slight blur in the reconstructed pictures are observed as Fig.\ref{fig:vaeloss}, which is inevitable but acceptable.

\section{Comparison with GaussianCube.}

\label{app:gaussian cube}
\begin{figure*}
    \centering
    \includegraphics[trim={1cm 2.9cm 16cm 2.5cm},clip, width=0.8\linewidth]{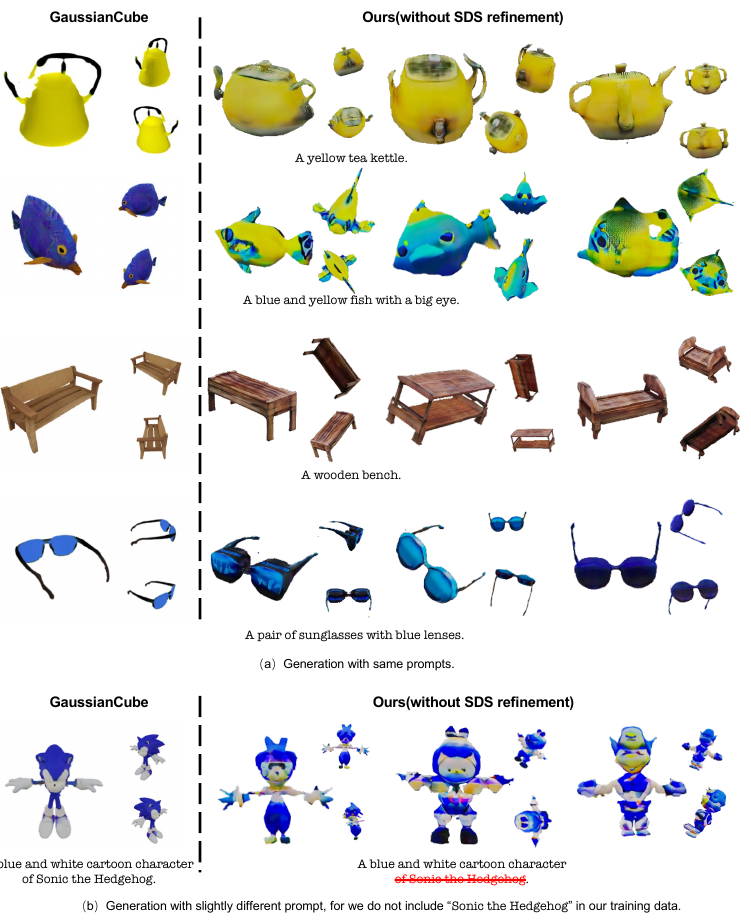}
    \caption{Comparison with GaussianCube. }
    \label{fig:gaussian_cube}
\end{figure*}

GaussianCube~\citep{zhang2024gaussiancube} is the most recent paper aiming to solve a similar task of ours, which can generate 3D GS directly without SDS or reconstruction from images.  As for now, the authors of  GaussianCube have not release their pre-trained models and runnable code for text-to-3D task on Objaverse dataset. Therefore, we just use the images provided in their paper for a qualitative comparison. The generated samples are shown in Fig.~\ref{fig:gaussian_cube}. Our method produces more diverse and detailing generation results.

\section{More Generation Results and Comparisons with LN3Diff and 3DTopia (without SDS Refinement).}
\begin{figure*}[b]
    \centering
    \includegraphics[trim={1cm 5cm 19cm 2.5cm},clip, width=0.9\linewidth]{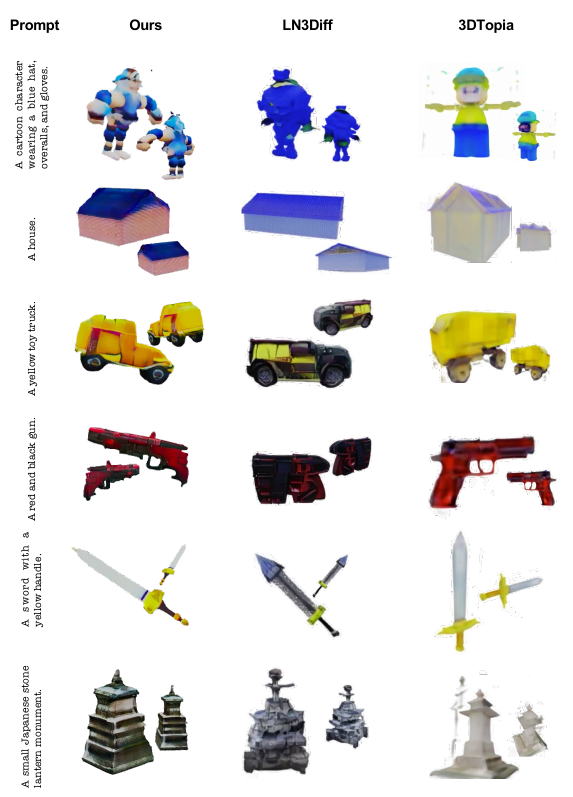}
    \caption{More generated samples (without SDS Refinement). }
    \label{fig:more1}
\end{figure*}

\begin{figure*}[b]
    \centering
    \includegraphics[trim={1cm 5cm 19cm 2.5cm},clip, width=0.9\linewidth]{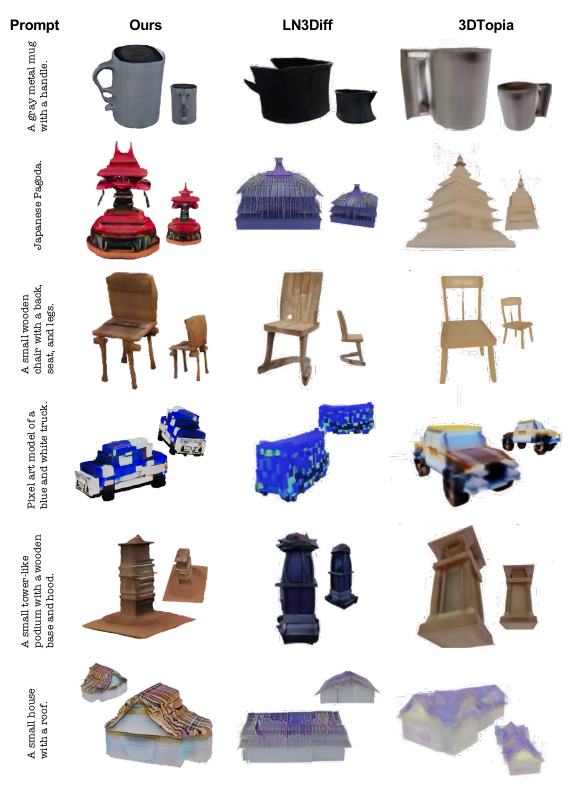}
    \caption{More generated samples compared with other methods(without SDS Refinement). }
    \label{fig:more4}
\end{figure*}

More generated samples compared with LN3Diff~\cite{lan2025ln3diff}, 3DTopia\cite{hong20243dtopia} and BrightDreamer\cite{jiang2024brightdreamer} are rendered as Fig.~\ref{fig:more1}$\sim$ Fig.~\ref{fig:more4}, where LN3Diff and 3DTopia are both direct generation methods exploiting Triplane as intermediate representation. The main differences between our work and these two alternatives are discussed as follows. 

First, while we select 3D GS as our 3D representation, LN3Diff and 3DTopia use NeRF as their target 3D representations, which unavoidably involves some disadvantages of neural field such as artifacts, implicitness for further editing, and slower rendering speed. As shown in these figures, our work enables a generation with explicit 3D GS, and fast rendering with a minimum of artifacts, especially compared with 3DTopia.

Second, we use explicit mesh surface for GS point binding, which enables more accurate texture projection. Moreover, different geometry losses help to constrain the mesh to be clear and smooth. Therefore, our method produces more detailing appearances without blurs compared with LN3Diff and 3DTopia, which can be further examined by zooming in the figures.

Moreover, there are several technical points are worth a discussion. For example, different from our end-to-end triplane VAE, LN3Diff adopts a image-to-triplane encoding. It is possible that the decoders cannot capture enough features from sparse input images for retrieving a high quality 3D object. Also, in such framework, different images sets of an unique object may corresponds to different latent codes, which is also a potential risk for following learning on latent space. As for 3DTopia, we guess that the absence of supervision for appearance detailing may the key reason for its blur output. In our method, the perceptual losses are incorporated in both the triplane encoding procedure and the VAE training.

\section{Comparison with BrightDreamer.}
\begin{figure*}[b]
    \centering
    \includegraphics[trim={1cm 11.5cm 19cm 2.5cm},clip, width=0.9\linewidth]{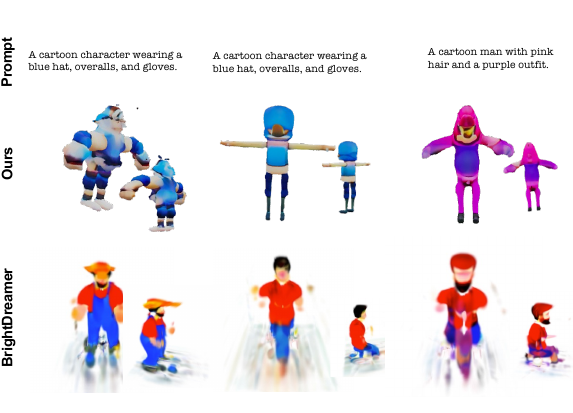}
    \caption{Case comparison with BrightDreamer. }
    \label{fig:more5}
\end{figure*}

BrightDreamer~\cite{jiang2024brightdreamer} is another methods that claim to generation 3D GS directly. However, it is trained on single class datasets created by Instant3D~\cite{li2024instant3d}, we cannot compare it with ours using various prompts as before. Therefore, we just pick several cases for demonstration. As shown in Fig.\ref{fig:more5}, while BrightDreamer also enables fast and direct GS generation, its output are more over-saturated, blurrier, and with more artifacts.


\begin{figure*}[b]
    \centering
    \includegraphics[trim={1cm 5cm 19cm 2.5cm},clip, width=0.9\linewidth]{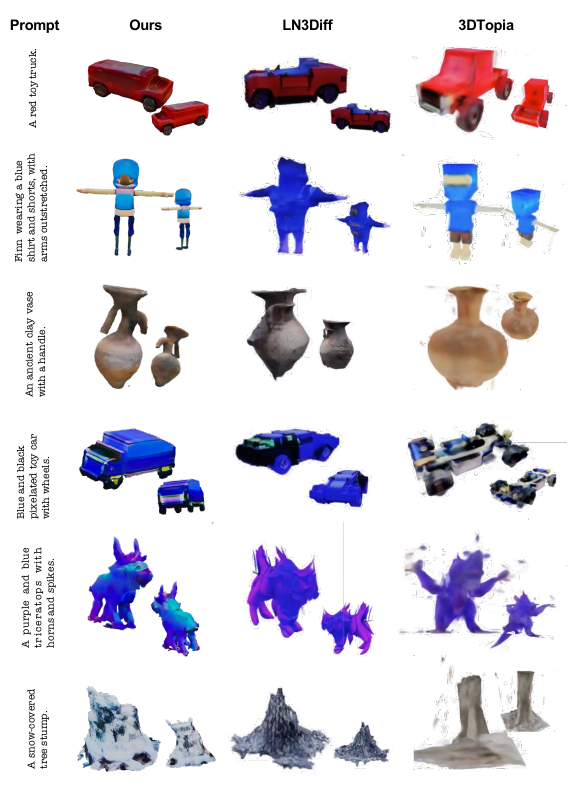}
    \caption{More generated samples (without SDS Refinement). }
    \label{fig:more2}
\end{figure*}

\begin{figure*}[b]
    \centering
    \includegraphics[trim={1cm 5cm 19cm 2.5cm},clip, width=0.9\linewidth]{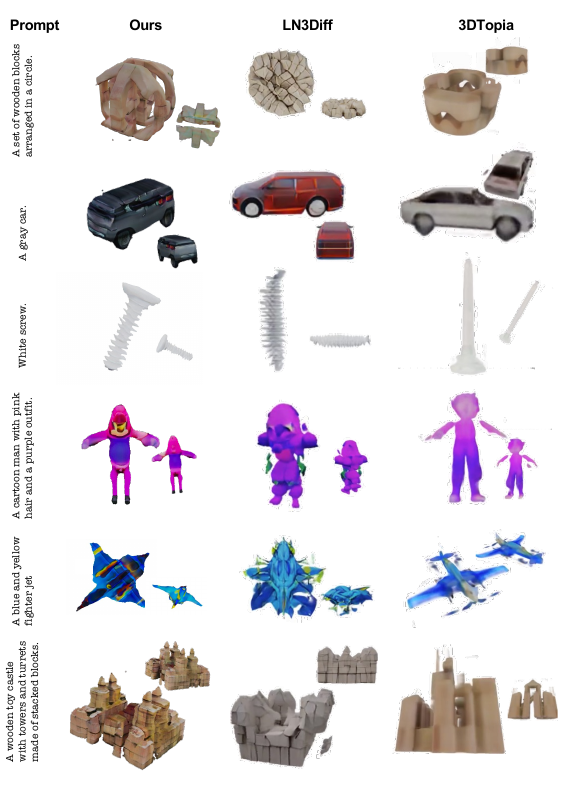}
    \caption{More generated samples compared with other methods (without SDS Refinement). }
    \label{fig:more3}
\end{figure*}

{
    \small
    \bibliographystyle{ieeenat_fullname}
    \bibliography{main}
}
\end{document}